\documentclass[runningheads]{llncs}
\usepackage{graphicx}
\usepackage{kotex}
\usepackage{mathtools}
\usepackage{multirow}
\usepackage{subfigure}
\graphicspath{ {images/} }
\usepackage{array}
\usepackage{cite}

\makeatletter
\newcommand{\thickhline}{%
    \noalign {\ifnum 0=`}\fi \hrule height 1pt
    \futurelet \reserved@a \@xhline
}
\newcolumntype{"}{@{\hskip\tabcolsep\vrule width 1pt\hskip\tabcolsep}}
\makeatother

\begin{document}
\title{Domain Generalization with Pseudo-Domain Label for Face Anti-Spoofing}
%
%
\author{Young Eun Kim\and Seong-Whan Lee}
\authorrunning{Y.E. Kim et al.}
%
\institute{Department of Artificial Intelligence \\
Korea University, Seoul, Republic of Korea \\
\email{\{ye\_kim, sw.lee\}@korea.ac.kr}}
\maketitle              
%
\begin{abstract}
Face anti-spoofing (FAS) plays an important role in protecting face recognition systems from face representation attacks.
Many recent studies in FAS have approached this problem with domain generalization technique.
Domain generalization aims to increase generalization performance to better detect various types of attacks and unseen attacks.
However, previous studies in this area have defined each domain simply as an anti-spoofing datasets and focused on developing learning techniques.
In this paper, we proposed a method that enables network to judge its domain by itself with the clustered convolutional feature statistics from intermediate layers of the network, without labeling domains as datasets.
We obtained pseudo-domain labels by not only using the network extracting features, but also using depth estimators, which were previously used only as an auxiliary task in FAS.
In our experiments, we trained with three datasets and evaluated the performance with the remaining one dataset to demonstrate the effectiveness of the proposed method by conducting a total of four sets of experiments.
\end{abstract}
\section{Introduction}

As face recognition (FR) has been widely studied over the past few decades \cite{fr1, fr2, fr3, fr4, fr5}, face recognition techniques have been applied to many real-word applications, such as surveillance and biometric systems. 
Unlike other biometric authentication systems, face-to-face authentication does not require expensive equipment and is also possible with regular RGB cameras.
However, FR systems are vulnerable to presentation attacks \cite{oulu, casia, replay, msu}.
Presentation attack is an attack in which someone attempts to authenticate using paper photos or videos of a face of other person.
Typical presentation attacks include printed photo attacks, replay attacks that play videos in front of authentication cameras using tablets or laptops, and 3d mask attacks, but various kinds of unseen attacks have recently been made to breach the security.
Thus, needs for technique to detect face presentation attacks has recently emerged and actively studied in the name of face anti-spoofing (FAS) for the security of authentication systems.

Prior work on FAS can be largely divided into two methods: temporal-based methods and texture-based methods.
In temporal based method, early works used particular liveness facial motions, such as mouth motion \cite{temporal1} and eye-blinking \cite{temporal2, temporal3} as spoofing cues. 
However, temporal-based methods is vulnerable to the attack of which an attacker punctures only the eyes or mouth of a photograph and allows the actual eye and mouth to replace it. The Texture based method extracts handcrafted features such as Local Binary Pattern (LBP) or Histogram of oriented Gradients (HoG) and uses traditional classifiers such as support vector machine (SVM) or linear discriminant analysis (LDA) to classify them to genuine face and spoof face \cite{texture1, texture2}. 
Texture based methods aim to learn the fine textures shown when a picture is printed and to learn the textures of the devices used for attack. 
Furthermore, beyond the methods using handcrafted features and traditional classifiers, deep learning-based methods were first proposed with convolutional neural networks \cite{CNN1}. 
In \cite{CNN4}, pixel-wise labeling was proposed demonstrating even better performance. 
However, while these existing studies have shown good performance on intra-dataset testing, they have shown significant degradation of performance on the cross-dataset testing. 
This means that existing methods is not robust on unseen attacks.

Research to improve the generalization performance of data of unseen domains has also evolved, called domain generalization (DG)\cite{dg1, dg2, dg3, dg4}. 
In FAS, studies have also shown to learn domain-invariant features by adapting several domain generalization techniques. 
Techniques such as adversarial learning, feature disentangling, and meta-learning were widely used to improve generalization performance. 
Adversarial learning \cite{adv} uses minmax games to make domain indistinguishable features using domain discriminator. 
For example, \cite{ssdg}, one discriminator and triplet loss were used together.
Feature division performs tasks by focusing on the spoof-relevant part, assuming that features extracted from the network can be distinguished between those containing a lot of domain-relevant information and those containing spoof-relevant information. For the first time in the \cite{rfmetafas}, the technique of metal learning for domain generalization was applied to anti-spoofing by using depth regression as an auxiliary task.

However, previous DG-focused studies have all defined each domain as dataset and focused on learning techniques. This is because each dataset has a different device or filming environment used in the attack. However, each dataset also contains different types of attacks, such as high quality attacks and low quality attacks or printed attacks and video replay attacks. The more datasets are used, the more likely they are to be included in other datasets, but the more similar types of attacks are to be classified as the same domain.

\begin{figure}[ht]
\centering
\subfigure[]{
    \includegraphics[width=2.7cm,height=3.2cm]{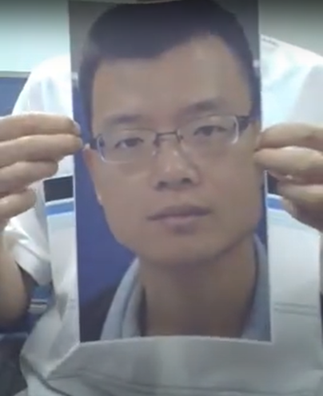}
    \label{fig:subfig1}
}
\subfigure[]{
    \includegraphics[width=2.7cm,height=3.2cm]{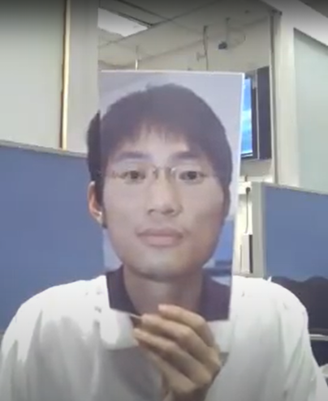}
    \label{fig:subfig2}
}
\subfigure[]{
    \includegraphics[width=2.7cm,height=3.2cm]{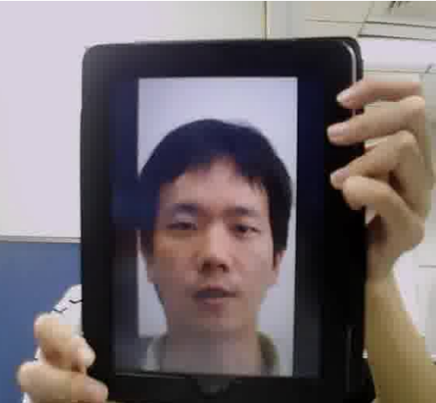}
    \label{fig:subfig3}
}
\subfigure[]{
    \includegraphics[width=2.7cm,height=3.2cm]{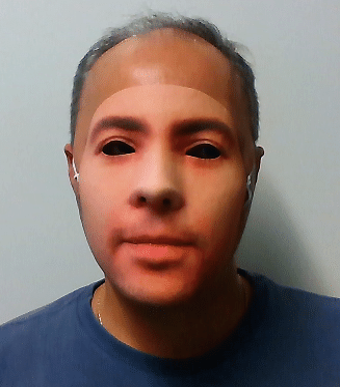}
    \label{fig:subfig4}
}
\caption[Optional caption for list of figures]{Various types of face presentation attacks \subref{fig:subfig1} printed photo attack, \subref{fig:subfig2} cut photo attack, \subref{fig:subfig3} video replay attack and \subref{fig:subfig4} 3D mask attack}
\label{fig:subfigureExample}
\end{figure}

In this paper, we do not simply define dataset as domain, but we propose to define domain by checking the style features of each dataset through output in the middle layer of the network, based on which we cluster. The \cite{style} states that the late feature of the middle layer of the network reflects the style information of the image, and the \cite{dgmmld} applies the late feature and adversarial learning of the middle layer to the domain generalization task. However, adversarial learning has the disadvantage of being unstable in learning, and setting different number of clustering groups can change the structure of the discriminator network, which can lead to performance degradation. Using this motif, we apply meta-learning techniques to extract domain-invited and spoof-relevent features from the network with a defined pseudo domain label. This provides stable learning, as well as the need to label domain if dataset is obtained by web-crawling.

\section{Related work}
\label{gen_inst}

\subsection{DG based Face anti-spoofing methods}
Domain generalization technique was widely used to improve the generalization performance of anti-spoofing algorithms against unsen attacks. 
Studies using Adversarial learning to extract domain-invariant features include \cite{ssdg} and \cite{maddog}. \cite{maddog} used as discriminators of which the number was same with the number of domains to allow the network to create domain-invariant features with a minimax approach. 
In \cite{ssdg}, the framework named SSDG is designed to generate domain invariant features by one discriminator. 
By passing adversarial learning only on the real image, the real image allows feature distribution to be distributed similarly regardless of its domains. 
\cite{disentangle} synthesized images by mixing the content features and livness features. 
By exchanging the liveness features of the real person and the attack, \cite{disentangle} got different reconstructed images with the same content but their liveness attributes are changed. 

Previous studies using meta learning include \cite{rfmetafas} and \cite{selfda}. Most similar to our study is the framework of \cite{rfmetafas}, which has a distribution that first proposed domain generalization metal learning to anti-spoofing. \cite{selfda} proposed an unsupplied domain adaptation method. Autoencoder is used to obtain domain information of data at the time of inference, and binary classification is used by classifiers based on domain information.
Among those works, the ones most related is \cite{rfmetafas}. 
The similarity is that we also used meta learning for domain generalization. 
However, we have significant difference in the aspect of defining domains. 
With our proposed method, the number of domain groups can be defined as much as we want.


\subsection{Meta learning for domain generalization (MLDG)}
Meta learning presents a technique that makes it applicable to any model rather than creating a new deep learning model for domain generalization performance. Learning to generalize describes the analysis of MLDG.
The objective of MLDG is:
\begin{equation}
\label{eq1}
\underset{\Theta}{\operatorname{argmin}} \mathcal{F}(\Theta)+\beta \mathcal{G}\left(\Theta-\alpha \mathcal{F}^{\prime}(\Theta)\right), 
\end{equation}
where $\mathcal{F}(\cdot)$ is the loss from the meta-train domains,  $\mathcal{G}(\cdot)$ is the loss from the meta-test domains and $\mathcal{F}^{\prime}(\Theta)$ is the gradient of the training loss $\mathcal{F}(\Theta)$.
It means meta learning will tune such that after updating the meta-train domains, performance is also good on the meta-test domains.

For another perspective view that \cite{ml4} is describing is that it we can get another analysis by doing the first order Taylor expansion for the second term in Eq. \ref{eq1} as follows:
\begin{equation}
\label{taylor}
\mathcal{G}(x)=\mathcal{G}(\dot{x})+\mathcal{G}^{\prime}(\dot{x}) \times(x-\dot{x}),
\end{equation}
where $\dot{x}$ is an arbitrary point that is close to $x$. The multi-variable form $x$ is a vector and $\mathcal{G}(x)$ is a scalar.

Using Taylor expansion in Eq. \ref{taylor}, Eq. \ref{eq1} can be modified as follows:
\begin{equation}
\underset{\Theta}{\operatorname{argmin}} \mathcal{F}(\Theta)+\beta \mathcal{G}(\Theta)-\beta \alpha\left(\mathcal{G}^{\prime}(\Theta) \cdot \mathcal{F}^{\prime}(\Theta)\right)
\end{equation}

This reveals that we want to minimize the loss on both meta-train and meta-test domains, and maximize the dot product of $\mathcal{F}^{\prime}(\Theta)$ and $\mathcal{G}^{\prime}(\Theta)$. 
Recall the dot operation computes the similarity of two vectors.
Since $\mathcal{F}^{\prime}(\Theta)$ and $\mathcal{G}^{\prime}(\Theta)$ are gradient of losses in two sets of domains, the similar direction means the direction of improvement in each set of domains is similar.

The DGML strategy in \cite{ml4} is used during training stage using pseudo-domain labels which were assigned to training data.


\section{Method}
\label{headings}
\begin{figure}
\centering
\label{fig2}
\includegraphics[width=12cm]{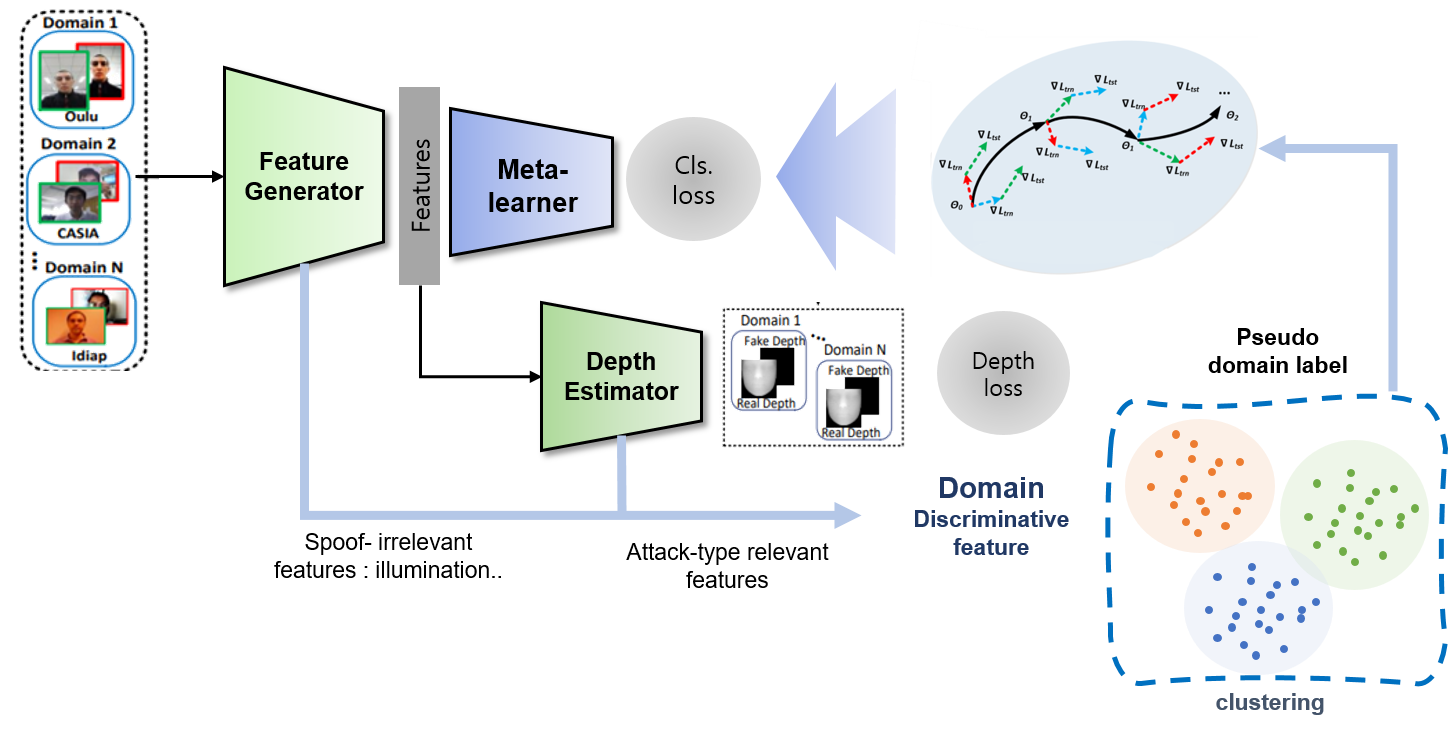}
\caption{Overview of the proposed framework. The stacked convolutional feature statistics from feature generator and depth estimator is clustered into $N$ groups and assigned pseudo-domain labels}
\end{figure}
\subsection{Overview}
In order to split domains of data without domain labels, we clustered the style features of each data iteratively.
Our method can be divided into two main processes. 
First, for every epoch, data are divided into N domains using the mean and variance of the mid-level features of the feature extractor and mid-level features of the depth estimator. 
Second, model-agnostic meta-learning (MAML) is exploited to generalize on domains.
The framework of our method is illustrated in Figure 1.

\subsection{Splitting domains}
Existing DG-based anti-spoofing studies define domains as datasets. 
However, if dataset are obtained through web Crawling, it is difficult to know the domain label and time is spent labeling it. 
We received a motivation from the previous study, \cite{dgmmld}, and use middle-level feature of the model to divide in respect to the domain of data. 
To define a domain, the middle-level feature is clustered with clustering model and labeled with N domains. 
The reason for using middle-level features is that in previous studies, middle-layer features are known to reflect the style of data \cite{style}.
We stacked the convolutional feature statistics of $5$th layer, $9$th layer of the feature extractor and the last layer of depth estimator, respectively. 
Then we adopted principal component analysis (PCA) to reduce the dimension of the stacked convolutional features into $256$ dimensions.
With the reduced style feature, clustering method like K-nearest neighbor (KNN) or gaussisan mixture model (GMM) was used to cluster source domain into $N$ pseudo domains.

\subsection{Meta learning for domain generalization}
In domain generalization, meta-learning performs three main steps repeatedly: meta-train, meta-test, and meta-optimization. 
Before performing meta-learning step, $(N-1)$ domains are randomly selected, and the remaining $1$ domain is defined as a meta-test set. 
In the meta-train phase, $(N-1)$ sets of meta-parameters obtained by performing 1 step gradient description based on each loss for $(N-1)$ datasets are obtained. 
Then, we use the meta-parameter of each domain to obtain the loss of data from the meta-test domain and perform the full optimization.

\begin{figure}
\label{fig1}
\centering
\includegraphics[width=12cm]{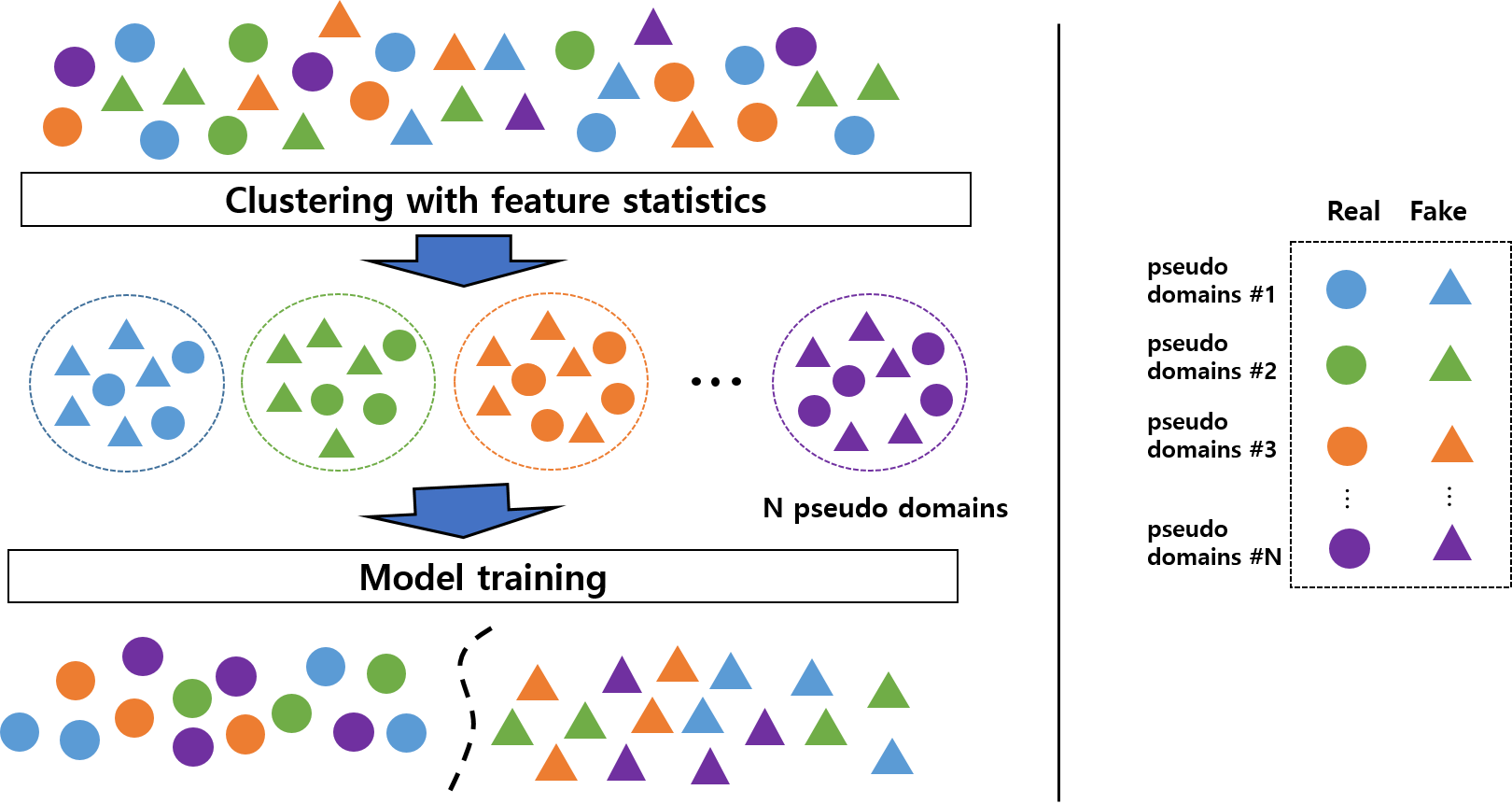}
\caption{Domain generalization with clustered convolutional feature statistics containing style information. Training data are clustered into $N$ pseudo-domain groups and trained in a way that classifies liveness well.}
\end{figure}

\paragraph{Meta-Train.}
We sampled batches in every meta-train domain $D_{\text {trn}}$, denoted as $\widehat{\mathcal{S}}_{i}(i=1, \ldots, N-1)$, and we conducted the cross-entropy classification and depth regression in every meta-train domain.
Cross entropy loss is calculated as follows:

\begin{equation}
\begin{aligned}
&\mathcal{L}_{Cls\left(\widehat{\mathcal{S}}_{i}\right)}\left(\theta_{F}, \theta_{M}\right) =\sum_{(x, y) \sim \widehat{\mathcal{S}}_{i}} y \log M(F(x))+(1-y) \log (1-M(F(x))),
\end{aligned}
\end{equation}
where $F$ and $M$ denote feature extractor and meta learner, respectively. $\theta_{F}$ is the parameters of the feature extractor, and $\theta_{M}$ is the parameters of the meta learner.
In each meta-train domain, we can obtain the updated meta learner's parameter, meta-parameters as $\theta_{M_{i}}{ }^{\prime}=\theta_{M}-\alpha \nabla_{\theta_{M}} \mathcal{L}_{C l s\left(\widehat{\mathcal{S}}_{i}\right)}\left(\theta_{F}, \theta_{M}\right)$.

Meanwhile, loss for depth estimator is calculated as follows:

\begin{equation}
\mathcal{L}_{\operatorname{Dep}\left(\widehat{\mathcal{S}}_{i}\right)}\left(\theta_{F}, \theta_{D}\right)=\sum_{(x, I) \sim \widehat{\mathcal{S}}_{i}}\|D(F(x))-I\|^{2},
\end{equation}
where $D$ denotes depth estimator, $\theta_{D}$ is the parameter of the depth estimator and $I$ are the pre-calculated face depth maps for input face images. 
We used the dense face alignment network named PRNet \cite{prnet} to obtain depth maps of real faces, and depth maps of all zeros are set as the supervision for fake faces.
Depth loss in meta train step is not used to update meta parameters, but used in meta optimization step.

\paragraph{Meta-Test.}
In Meta-test phase, we sampled batch in the one remaining meta-test domain $D_{val}$, denoted as $\tilde{\mathcal{S}}$.
To generalize well to unseen attacks of various senarios, we use meta-parameters which was obtained from meta-train phase to calculate the classification losses:
\begin{equation}
\begin{aligned}
&\sum_{i=1}^{N-1} \mathcal{L}_{Cls(\widetilde{\mathcal{S}})}\left(\theta_{F}, \theta_{M_{i}}^{\prime}\right) =
&\sum_{i=1}^{N-1} \sum_{(x, y) \sim \tilde{\mathcal{S}}} y \log M_{i}^{\prime}(F(x))+(1-y) \log \left(1-M_{i}^{\prime}(F(x))\right)
\end{aligned}
\end{equation}

The depth loss is also calculated like meta-train:
\begin{equation}
\mathcal{L}_{\operatorname{Dep}(\widetilde{\mathcal{S}})}\left(\theta_{F}, \theta_{D}\right)=\sum_{(x, I) \sim \widetilde{S}}\|D(F(x))-I\|^{2}
\end{equation}

By using meta parameters in calculating classification loss on meta-test domain, we can obtain the generalized direction of gradients of meta-test loss.

\paragraph{Meta-optimization.}
To update the parameter of feature extractor, meta learner and depth estimator, we summarize all the learning information in the meta-train and meta-test as follows:

\begin{equation}
\theta_{M} \leftarrow \theta_{M}-\beta \nabla_{\theta_{M}}\left(\sum_{i=1}^{N-1}\left(\mathcal{L}_{C l s\left(\widehat{\mathcal{S}}_{i}\right)}\left(\theta_{F}, \theta_{M}\right)+\mathcal{L}_{C l s(\widetilde{\mathcal{S}})}\left(\theta_{F}, \theta_{M_{i}}^{\prime}\right)\right)\right)
\end{equation}

\begin{equation}
\begin{aligned}
\theta_{F} \leftarrow \theta_{F}-\beta \nabla_{\theta_{F}} &\left(\mathcal{L}_{\operatorname{Dep}(\tilde{\mathcal{S}})}\left(\theta_{F}, \theta_{D}\right)+\sum_{i=1}^{N-1}\left(\mathcal{L}_{Cls \left(\widehat{\mathcal{S}}_{i}\right)}\left(\theta_{F}, \theta_{M}\right)\right.\right. \\
&\left.\left.+\mathcal{L}_{Dep\left(\widehat{\mathcal{S}}_{i}\right)}\left(\theta_{F}, \theta_{D}\right)+\mathcal{L}_{Cls(\widetilde{\mathcal{S}})}\left(\theta_{F}, \theta_{M_{i}}^{\prime}\right)\right)\right)
\end{aligned}
\end{equation}

By doing meta-train, meta-test and meta-optimization stage iteratively, we can extract generalized feature which are related to spoof discriminative information and discard spoof-irrelevant information.
\section{Experimental evaluation}
\label{others}
\subsection{Experimental settings}
\paragraph{Datasets.} Four public face anti-spoofing datasets are utilized to evaluate the effectiveness of our method: OULU-NPU \cite{oulu}, CASIA-MFSD \cite{casia}, Idiap Replay-Attack \cite{replay}, and MSU-MFSD \cite{msu}.
Following the setting in \cite{rfmetafas}, one dataset is treated as one domain in our experiment.
we randomly selected three datasets as source domains for training and the remaining one as the target domain for testing.
Thus, we have four testing tasks in total: O\&C\&I to M, O\&M\&I to C, O\&C\&M to I, and I\&C\&M to O.
Area under curve (AUC) is used as the evaluation metric.
Also, t-SNE visualization is also reported to further evaluate performance.

\begin{table}[]
\caption{Comparison to face anti-spoofing methods on four testing sets for domain generalization on face anti-spoofing}
\label{extable}
\centering
\resizebox{\textwidth}{!}{
\begin{tabular}{l|c|c|c|c}
\thickhline
\multicolumn{1}{c}{\multirow{2}{*}{Methods}} & \textbf{O\&C\&I to M}  & \textbf{O\&M\&I to C}  & \textbf{O\&C\&M to I}  & \textbf{I\&C\&M to O}  \\ \cline{2-5} 
\multicolumn{1}{c|}{}                         & AUC(\%)       & AUC(\%)       & AUC(\%)       & AUC(\%)       \\ \thickhline
MS\_LBP                         & 78.5          & 44.9          & 61.6          & 49.3          \\ \hline
Binary CNN (ICB 2019)                         & 82.8          & 71.9          & 65.8          & 77.5          \\ \hline
Auxiliary(Depth)                              & 85.8          & 73.1          & 75.1          & 77.6          \\ \hline
MADDG (AAAI 2019)                             & 88            & 84.5          & 84.9          & 80            \\ \hline
RFM (AAAI 2020)                               & \textbf{93.9} & \textbf{88.1} & \textbf{90.4} & \textbf{91.1} \\ \hline
Self-DA (AAAI 2021)                           & 91.8          & 84.4          & 90.1          & 84.3          \\ \thickhline
PDL-FAS (proposed)                       & 93.1          & 87.6          & 88.4          & 90.7          \\ \thickhline
\end{tabular}
}
\end{table}

\paragraph{Implementation details.} The size of face image is $256 \times 256 \times$ 6, where we extracted RGB and HSV channels of each face image.
The Adam optimizer is used for the optimization.
The learning rate $\alpha, \beta$ are set to 0.001. The batch size is 7 per domain, and thus 21 for 3 training domains totally.
For a new testing sample, its classification score $l$ is calculated for testing as follow: $l=M(F(x))$, where $F$ and $M$ are the trained feature extractor and meta learner.
The number of clustering group was set to three.

\subsection{Experimental comparison}
\paragraph{Compared methods.} We compare several state-of-the-art
face anti-spoofing methods as follows: 
Multi-Scale LBP \cite{mslbp}; 
Binary CNN \cite{binarycnn}; 
Auxiliary \cite{auxiliary}: This method learns a CNN-RNN model to estimate the face depth from one frame and rPPG signals through multiple frames.
To fairly compare our method only using one frame information, we also compare the results of its face depth estimation component (denoted as Auxiliary(Depth));
MADDG \cite{maddog}: This method uses adversarial learning for domain generalization;
RFM \cite{rfmetafas}: This method uses meta learning which is similar to our works;
SelfDA \cite{selfda}: This method is for domain adaptation, but similar with domain generalization in that it doesn't have target domain information.

\paragraph{Comparison Results with SOTA Face Anti-Spoofing
Methods}
From comparison results in Table \ref{extable}, it can be seen that the proposed method outperforms most of the state-of-the-art face anti-spoofing methods.
Moreover, as shown in Figure $4$, we randomly select 1000 samples of target dataset and plot the t-SNE visualizations to analyze the feature space learned by our proposed method.
Even though it didn't reach at the best result, this method has advantages that it doesn't require domain labels and can adjust the number of domain we are going to use in meta-learning.

\begin{figure}[ht]
\centering
\subfigure[]{
    \includegraphics[width=5.7cm]{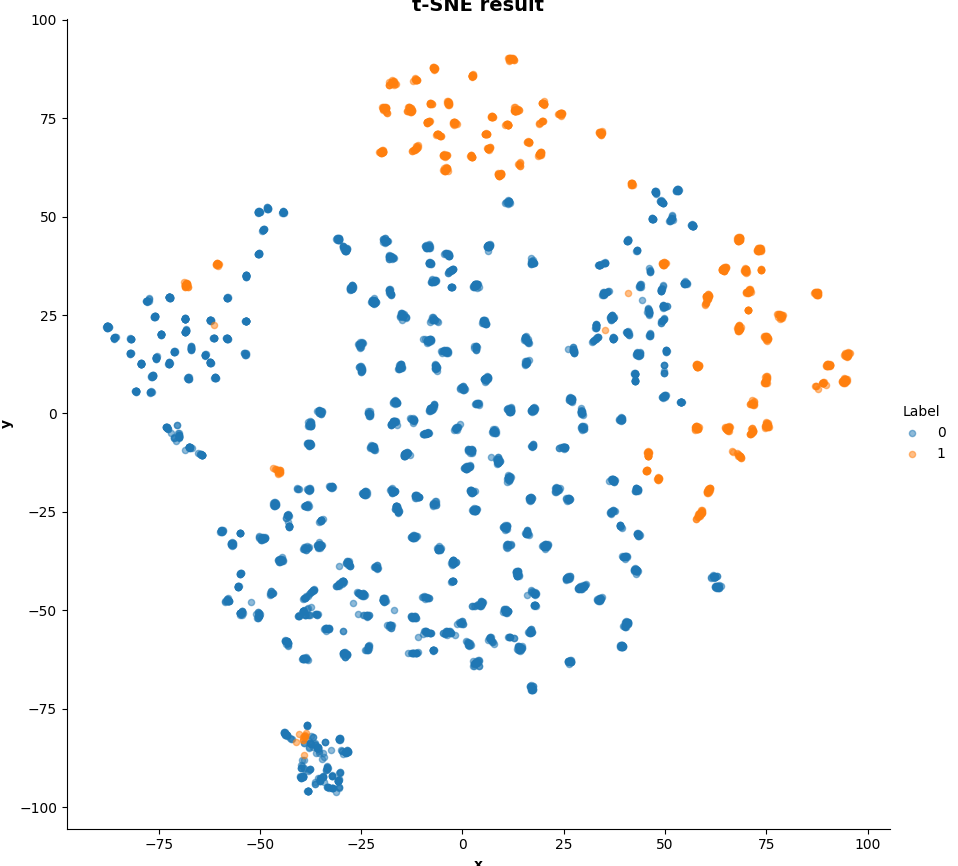}
    \label{fig:subfig1}
}
\subfigure[]{
    \includegraphics[width=5.7cm]{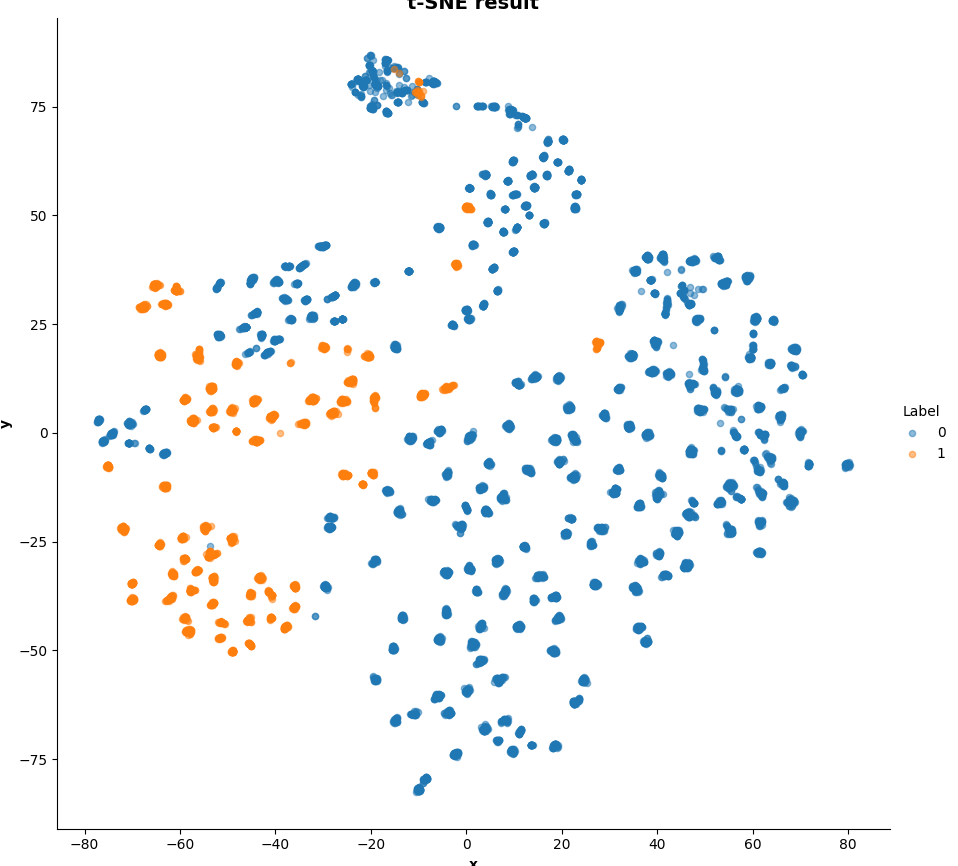}
    \label{fig:subfig2}
}

\caption[]{The t-SNE visualization of our model with assigning pseudo-domain labels. Blue colors are fake test images and orange colors are real images. \subref{fig:subfig1} t-SNE plot which is tested with MSU-MFSD and trained on Oulu-NPU, CASIA-FASD, Idiap Replay attack. \subref{fig:subfig2} t-SNE plot which is tested with CASIA-FASD and trained on Oulu-NPU, Idiap Replay attack, MSU-MFSD}
\label{fig:subfigureExample}
\end{figure}

\section{Conclusion}
In this paper, we cluster using the average and variance of the late features extracted from the feature extractor and the depth estimator, and define N pseudo-domains. 
It also applied meta learning with a dataset of psuedo domain defined for Domain generalization. 
In the meta train phase, each loss was obtained for a different domain and N meta-parameters for one step optimization, and eventually the meta test found a direction in which the loss was minimized for the new dataset. 
Although this study did not show high performance compared to the method of SOTA, it has the advantage of not requiring domain labeling unlike conventional methods.
Also, domain can be defined separately by the number of domains desired.

\end{document}